\documentclass[sigconf]{acmart}
\AtBeginDocument{%
  }

\setcopyright{acmlicensed}
\copyrightyear{2018}
\acmYear{2018}
\acmDOI{XXXXXXX.XXXXXXX}
\acmConference[Conference acronym 'XX]{Make sure to enter the correct
  conference title from your rights confirmation email}{June 03--05,
  2018}{Woodstock, NY}

\acmISBN{978-1-4503-XXXX-X/2018/06}
\usepackage{graphicx}
\usepackage{subfigure}
\usepackage{amsmath}
\usepackage{amsthm}
\usepackage{booktabs}
\usepackage{algpseudocode,algorithm}
\usepackage[switch]{lineno}
\usepackage{enumitem}
\usepackage{multirow}

\begin{document}

\title{Reasoning While Recommending: Entropy-Guided Latent Reasoning in Generative Re-ranking Models}

\author{Changshuo Zhang}
\affiliation{%
  \institution{Individual Author}
  % \city{Beijing}
  \country{China}
}
\email{lyingCS@foxmail.com}

\renewcommand{\shortauthors}{Changshuo Zhang et al.}
\newcommand{\ourname}{EGLR}

\begin{abstract}

Reinforcement learning plays a crucial role in generative re-ranking scenarios due to its exploration-exploitation capabilities, but existing generative methods mostly fail to adapt to the dynamic entropy changes in model difficulty during list generation, making it challenging to accurately capture complex preferences. Given that language models have achieved remarkable breakthroughs by integrating reasoning capabilities, we draw on this approach to introduce a latent reasoning mechanism, and experimental validation demonstrates that this mechanism effectively reduces entropy in the model’s decision-making process. 
Based on these findings, we introduce the \textbf{E}ntropy-\textbf{G}uided \textbf{L}atent \textbf{R}easoning~(\ourname) recommendation model, which has three core advantages. First, it abandons the "reason first, recommend later" paradigm to achieve "reasoning while recommending", specifically designed for the high-difficulty nature of list generation by enabling real-time reasoning during generation. Second, it implements entropy-guided variable-length reasoning using context-aware reasoning token alongside dynamic temperature adjustment, expanding exploration breadth in reasoning and boosting exploitation precision in recommending to achieve a more precisely adapted exploration-exploitation trade-off. Third, the model adopts a lightweight integration design with no complex independent modules or post-processing, enabling easy adaptation to existing models. Experimental results on two real-world datasets validate the model’s effectiveness, and its notable advantage lies in being compatible with existing generative re-ranking models to enhance their performance. Further analyses also demonstrate its practical deployment value and research potential.
% The code is available at~\textcolor{blue}{\url{https://anonymous.4open.science/r/EGLR-01C8}}.
\end{abstract}




\maketitle

\section{Introduction}\label{sec:intro}
\begin{figure}[tbp]  % [htbp]: Controls floating priority (here, top, bottom, page)
    \centering  % Center-align the image
    \includegraphics[width=1.0\linewidth]{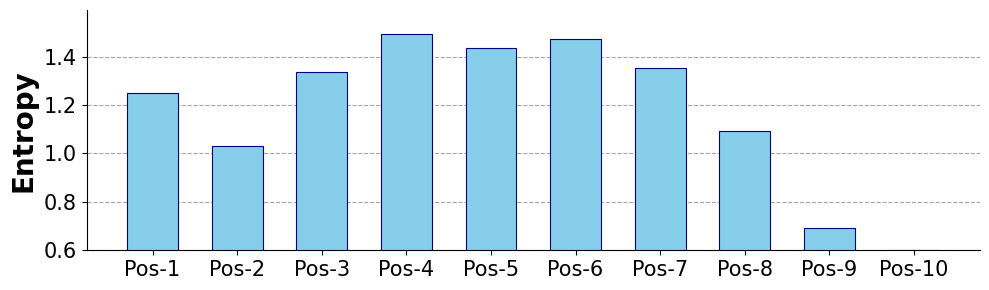}  
    \vspace{-0.6cm}
    \caption{Entropy change of the generative re-ranking model on different position during auto-regressive generation.}
    \vspace{-0.2cm}
    \label{fig:ent_GRM}
\end{figure}

\begin{figure}[tbp]  % [htbp]: Controls floating priority (here, top, bottom, page)
    \centering  % Center-align the image
    \includegraphics[width=1.0\linewidth]{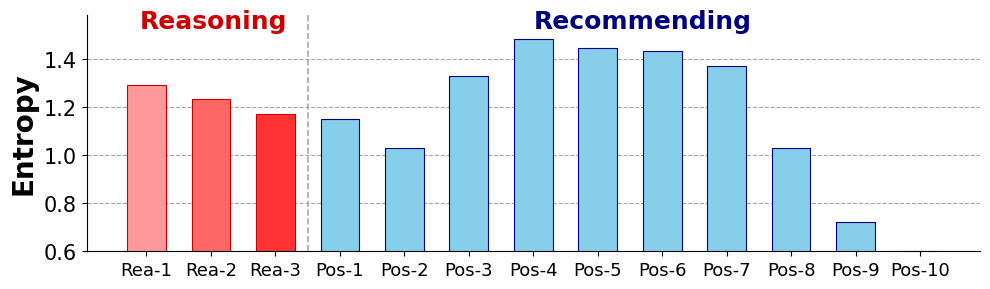}  
    \vspace{-0.6cm}
    \caption{Entropy change of the generative re-ranking model with latent reasoning during auto-regressive generation.}
    \vspace{-0.2cm}
    \label{fig:ent_rea}
\end{figure}

Recommender systems (RS) are indispensable components of modern real-world platforms, with wide applications across domains such as short-video~\citep{zhan2022deconfounding, zhao2024counteracting, zhang2025comment}, e-commerce~\citep{pei2019personalized, wang2024not, hussien2021recommendation}, music streaming~\citep{zhang2022counteracting, bendada2020carousel, van2013deep}, news delivery~\citep{wu2023personalized, ozgobek2014survey, liu2024news}, live streaming~\cite{rappaz2021recommendation, gao2023live, qu2025kuailive}, and knowledge sharing~\citep{shen2009recommending, zhang2024qagcf}. Generative Re-ranking Models (GRM) models lie in the final re-ranking stage of the recommendation pipeline: they leverage outputs from upstream components, model inter-item dependencies within the candidate pool, and directly determine the final recommended list~\cite{huzhang2021aliexpress, chen2023controllable, wang2024not}.  

Recent GRM adopt the Generator-Evaluator architecture, where the Generator harnesses the exploration-exploitation capabilities of reinforcement learning (RL)~\cite{schulman2017proximal, sutton1998reinforcement} to learn optimal list generation strategies by interacting with the Evaluator (i.e., the reward model). However, the exploration-exploitation potential of RL remains underexplored in GRM. As illustrated in Figure~\ref{fig:ent_GRM}, we use the entropy of the candidate set to quantify the task difficulty during the generation process of the well-trained CMR~\cite{chen2023controllable} model on the Ad\footnote{https://tianchi.aliyun.com/dataset/56} dataset and find that entropy exhibits a non-uniform distribution: it is low in the initial and final stages of list generation but peaks in the middle segment. This distribution pattern stems from dense positive feedback signals and clear reward cues in the early stage, coupled with a shrinking candidate pool in the late stage; by contrast, the middle segment involves a larger candidate space and more frequent negative rewards, resulting in elevated entropy.

Building on recent advances that have successfully transplanted language models’ latent reasoning capabilities~\cite{zhu2025survey, xu2024lars, saunshi2024inductive, deng2024explicit, hao2024training} into recommender system to unlock inferential potential~\cite{zhang2025reinforced, zhang2025slow, tang2025think}, we integrate this latent reasoning mechanism into GRM, specifically by performing $3$ autoregressive steps before initiating generative re-ranking, and analyze entropy dynamics during the reasoning process. As shown in Figure~\ref{fig:ent_rea}, while latent reasoning generally reduces candidates selection entropy over time, entropy still surges in the middle of list generation—indicating that although the model can mitigate overall decision-making entropy through latent reasoning, it fails to adequately address the high-difficulty middle phase of list generation. 

Guided by this insight, we propose the \textbf{E}ntropy-\textbf{G}uided \textbf{L}atent \textbf{R}easoning (\ourname) recommendation model, with three key contributions: (1) Targeting the high-difficulty middle phase of list generation where GRM’s core challenge lies, \ourname~abandons the inadequate "reason first, recommend later" paradigm and adopts a "reasoning while recommending" framework to enable real-time reasoning during generation. (2) It integrates entropy-guided context-aware reasoning and dynamic temperature coefficient adjustment by expanding exploration breadth in reasoning and boosting exploitation precision in recommendation to synergistically balance exploration and exploitation in RL-driven GRM. (3) \ourname~features a lightweight integration design with no complex independent modules or post-processing to enable easy adaptation to existing generative re-ranking models.

The major contributions of the paper are as follows:

\begin{itemize}[leftmargin=*]
\item We identified the non-uniform entropy distribution in generative re-ranking, revealing that task difficulty peaks in the middle of list generation. We further introduced latent reasoning into GRM and analyzed its entropy dynamics, demonstrating that while latent reasoning reduces overall decision entropy it fails to address the high-difficulty middle phase thereby laying the foundation for targeted optimization.
\item We proposed \ourname, which integrates three synergistic core designs: a "reasoning while recommending" paradigm for real-time reasoning, entropy-guided context-aware latent reasoning, and dynamic temperature adjustment that effectively addresses the exploration-exploitation imbalance in RL-driven GRM.
\item \ourname~adopts a lightweight non-intrusive integration design, enabling easy adaptation to existing generative re-ranking frameworks without redundant modifications. Experiments on two real-world datasets validate its effectiveness.
\end{itemize}
\section{Related Work}

\subsection{Generative Re-ranking Models}
Generative Re-ranking Models (GRMs) act as the final re-ranking stage component of multi-stage recommender systems~\cite{ren2024non, carraro2024enhancing, lin2024discrete, gao2024llm, han2023personalized}, operating after point-wise prediction models to refine candidate list order~\cite{ai2018learning, ai2019learning, zhuang2018globally, pang2020setrank, pei2019personalized}. They explicitly consider list-wise context, specifically the mutual influence between items. This feature sets them apart from upstream ranking stages. Re-ranking models are divided into two training paradigms: learning from observed signals (needing optimal list labels, challenging for large spaces) and counterfactual signals~\cite{liu2022neural}. GRMs belong to the latter, typically built on Generator-Evaluator frameworks. Pioneered by EG-Rerank using RL for exploration~\cite{huzhang2021aliexpress}, later works like CMR~\cite{chen2023controllable} and LAST~\cite{wang2024not} improved expressiveness. However, existing GRMs still underutilize RL’s exploration-exploitation potential and fail to address high-entropy middle segments in list generation.

\subsection{Latent Reasoning in Recommendation}
Recent progress in recommender systems has transplanted the inferential capabilities of language models~\cite{zhu2025survey, xu2024lars, saunshi2024inductive, deng2024explicit, hao2024training, saunshi2025reasoning, saunshi2025reasoning, wang2023guiding, gatmiry2024can, chen2025reasoning} via latent reasoning~\cite{zhang2025reinforced, tang2025think, zhang2025slow}. This approach aims to unlock multi-step decision-making potential beyond direct pattern matching, though most related works focus on sequential recommendation scenarios.
STREAM-Rec, for example, proposes a three-stage training framework to mitigate the limitations of one-step inference~\cite{zhang2025slow}. ReaRec enhances user representations through implicit multi-step reasoning, paired with lightweight refinement methods like ERL and PRL~\cite{tang2025think}. LatentR$^3$ adopts two-stage training, combining supervised fine-tuning and modified GRPO-based reinforcement learning to optimize recommendations without relying on explicit chain-of-thought data~\cite{zhang2025reinforced}.
Yet latent reasoning has rarely been integrated into generative re-ranking scenarios. Existing studies do not leverage its ability to reduce decision entropy, leaving the core challenges of GRM list generation unaddressed.

\begin{figure*}[tbp]  % htbp：控制图片浮动优先级（here, top, bottom, page）
    \centering
    \includegraphics[width=\linewidth]{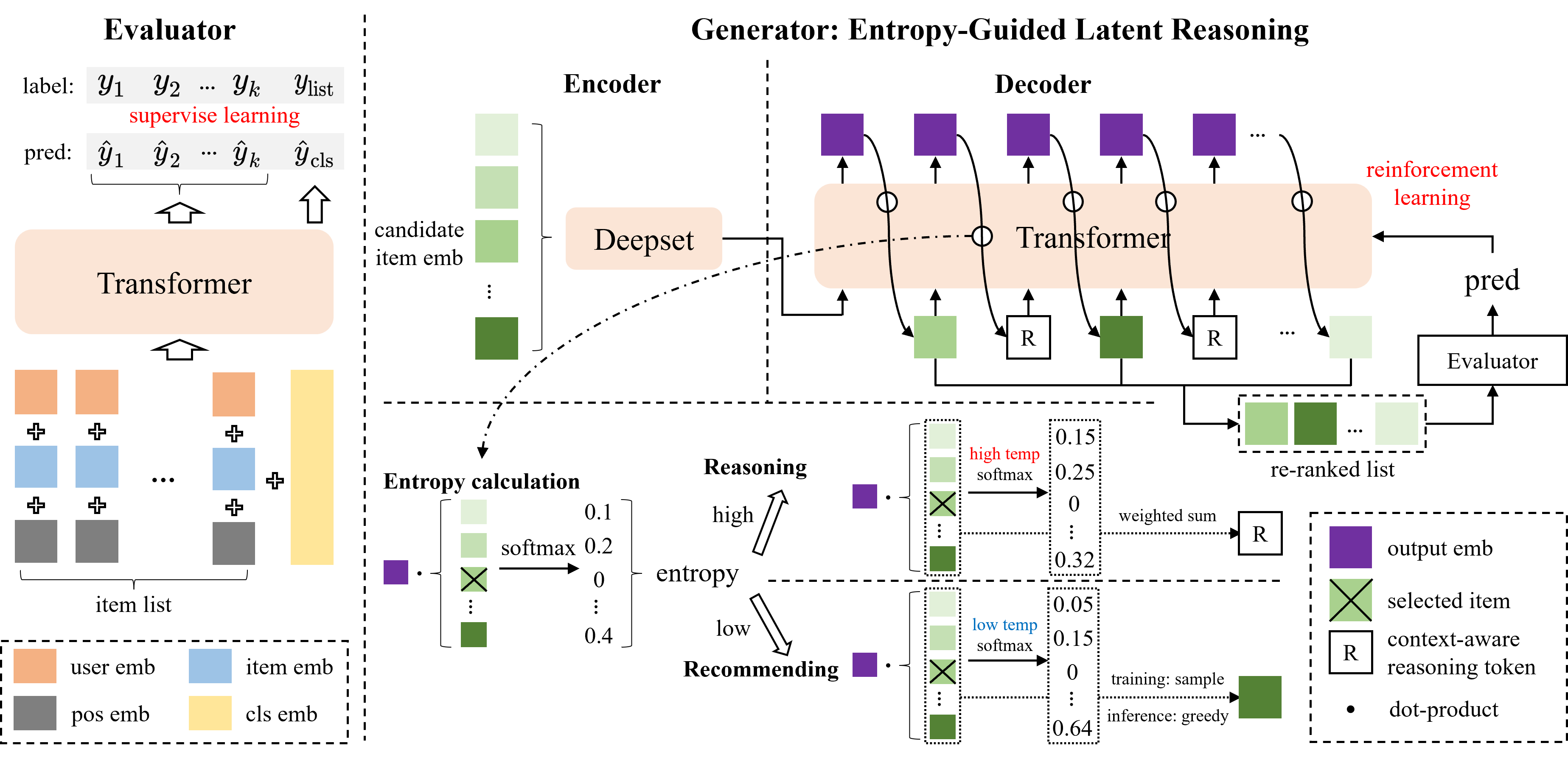}
    \caption{Architecture of \ourname: The left branch is the Evaluator, pre-trained on historical data and using a Transformer with dual heads to predict item-wise and list-wise feedback. The right branch is the Generator, consisting of an order-agnostic Encoder and an autoregressive Decoder. It incorporates entropy-guided context-aware reasoning tokens and dynamic temperature adjustment—high temperature for reasoning stages and low temperature for recommending stages. The Evaluator provides reward signals to update the Generator via reinforcement learning.}
    \label{fig:main}
\end{figure*}

\section{Problem Formulation}
We formally define the task of generative re-ranking and elaborate on the Generator-Evaluator framework adopted for model training.

Let \( \mathcal{U} \) denote the set of users, \( \mathcal{I} = \{i_1, i_2, \dots, i_M\} \) represent the candidate item pool (with \( M \) items), and \( K \) (where \( K \leq M \)) be the fixed length of the target recommended list. For a given user \( u \in \mathcal{U} \), the GLR task aims to generate an optimal ordered list \( \mathcal{L} = [l_1, l_2, \dots, l_K] \subseteq \mathcal{I} \) (where \( l_k \in \mathcal{I} \) is the \( k \)-th item in the list) that maximizes the user's overall satisfaction, considering both individual item preferences and inter-item dependencies (e.g., complementarity or redundancy).

To model this process, we employ a Generator-Evaluator architecture, consisting of two core components:

\noindent \textbf{Generator (\( \mathcal{G} \))}: A parametric model \( \mathcal{G}(\cdot; \theta) \) (with parameters \( \theta \)) responsible for list generation. It takes as input the user \( u \) and candidate pool \( \mathcal{I} \), and sequentially outputs items to form the list \( \mathcal{L} \) through a step-wise decision process:
\begin{equation}
\mathcal{L} = \mathcal{G}(u, \mathcal{I}; \theta) = \left[ \mathcal{G}_1(u, \mathcal{I}; \theta), \mathcal{G}_2(u, \mathcal{I}, l_1; \theta), \dots, \mathcal{G}_K(u, \mathcal{I}, l_1, \dots, l_{K-1}; \theta) \right],
\end{equation}
where \( \mathcal{G}_k(\cdot) \) denotes the generator's decision at step \( k \), conditioned on previously selected items \( l_1, \dots, l_{k-1} \).

\noindent \textbf{Evaluator (\( \mathcal{E} \))}: A reward model \( \mathcal{E}(\cdot; \phi) \) (with parameters \( \phi \)) that evaluates the quality of the generated list \( \mathcal{L} \). It outputs a list-wise reward score reflecting how well \( \mathcal{L} \) aligns with user preferences and platform objectives (e.g., engagement, diversity):
\begin{equation}
\hat{r} = \mathcal{E}(u, \mathcal{L}; \phi),
\end{equation}
where \( \hat{r} \) approximates the true user feedback \( r \) (e.g., cumulative clicks, conversion rate) for list \( \mathcal{L} \).

The training process of the Generator-Evaluator framework proceeds in two stages:

\paragraph{Evaluator Pre-training} The evaluator is first trained to approximate the true reward function using historical interaction data. Given a set of observed lists \( \{\mathcal{L}^{(t)}\}_{t=1}^T \) with their corresponding true rewards \( \{r^{(t)}\}_{t=1}^T \), the goal is to minimize the prediction error:
\begin{equation}
\min_{\phi} \mathbb{E}_{\mathcal{L}, r \sim \mathcal{D}} \left[ \mathcal{L}_{\text{eval}} \left( \mathcal{E}(u, \mathcal{L}; \phi), r \right) \right],
\end{equation}
where \( \mathcal{D} \) is the historical dataset, and \( \mathcal{L}_{\text{eval}}(\cdot, \cdot) \) is a loss function (e.g., cross-entropy for discrete feedback).

\paragraph{Generator Training} Guided by the pre-trained evaluator, the generator is optimized to maximize the expected reward of the generated lists. Its objective is:
\begin{equation}
\max_{\theta} \mathbb{E}_{u \sim \mathcal{U}, \mathcal{I} \sim \mathcal{I}_{\text{pool}}} \left[ \mathcal{E} \left( u, \mathcal{G}(u, \mathcal{I}; \theta); \phi \right) \right],
\end{equation}
where \( \mathcal{I}_{\text{pool}} \) denotes the distribution over candidate pools encountered in practice.

Notably, the generator's step-wise generation process introduces dynamic entropy, which we quantify using the entropy of the candidate set distribution at each step \( k \):
\begin{equation}
H_k = -\sum_{i \in \mathcal{I} \setminus \{l_1, \dots, l_{k-1}\}} p_k(i) \log p_k(i),
\end{equation}
where \( p_k(i) \) is the probability that item \( i \) is selected at step \( k \) by the generator. This entropy \( H_k \) serves as a key indicator of task difficulty during list generation, as analyzed in Section 1.

\section{\ourname: The Proposed Algorithm}

In this section, we will introduce our proposed model~\ourname.
\subsection{Overview of \ourname}
The overall framework of \ourname~is illustrated in Figure~\ref{fig:main}, consisting of three core components: (1) We build the backbone Generator-Evaluator framework where the Evaluator predicts user feedback for item lists via supervised learning, while the Generator processes all candidate items for re-ranking. (2) We equip the Generator with latent reasoning capability, achieving "reason while recommending" through real-time entropy monitoring and dynamic temperature coefficient adjustment during autoregressive generation. (3) The Evaluator is optimized via supervised learning using real user feedback, and the Generator is updated with GRPO based on rewards derived from the Evaluator’s scores for the re-ranked item list.

% \section{\ourname: The Proposed Algorithm}
% To address the high-entropy middle segment in generative re-ranking—an issue we empirically validated in Section 1—we propose \ourname~(Entropy-Guided Latent Reasoning). This model embeds entropy-aware latent reasoning into the standard Generator-Evaluator framework, with designs tailored to balance entropy reduction, exploration-exploitation (E\&E) dynamics, and real-world deployment efficiency. Below, we detail its core architecture, reasoning mechanisms, and training pipeline.

\subsection{Backbone Generator-Evaluator Architecture}
\ourname’s backbone relies on a Generator-Evaluator paradigm, where the Evaluator first learns to quantify list quality from historical data, and the Generator then optimizes list generation under this learned supervision. We prioritize detailing the Evaluator first, as its reward signals directly guide the Generator’s training.

\subsubsection{Evaluator Architecture}
The Evaluator \( \mathcal{E} \) takes an item list \( \mathcal{L} = [l_1, l_2, \dots, l_K] \) and a target user \( u \) as input, processing them to output both granular item feedback and holistic list quality. Its workflow starts with embedding construction that tightly integrates user features and positional information, followed by context aggregation via Transformer, and concludes with dual-output prediction.

For each item \( l_k \) in \( \mathcal{L} \), a base embedding \( \mathbf{e}_{l_k}^{\text{base}} \in \mathbb{R}^{d_{\text{item}}} \) (capturing content features like category or price) is first expanded by concatenating the user’s embedding \( \mathbf{e}_u \in \mathbb{R}^{d_{\text{user}}} \) along the feature dimension (second dimension), forming a user-item joint embedding:
\begin{equation}
\mathbf{e}_{l_k, u} = \mathbf{e}_{l_k}^{\text{base}} \oplus \mathbf{e}_u \in \mathbb{R}^{d}, \quad d = d_{\text{item}} + d_{\text{user}}.
\end{equation}
These joint embeddings form a list matrix \( \mathbf{E}_{\text{joint}} = [\mathbf{e}_{l_1, u}, \mathbf{e}_{l_2, u}, \dots, \mathbf{e}_{l_K, u}]^T \in \mathbb{R}^{K \times d} \). To model the order of items in the list, a sinusoidal position embedding $\mathbf{P} \in \mathbb{R}^{K \times d}$ is concatenated to \( \mathbf{E}_{\text{joint}} \):
\begin{equation}
\mathbf{E}_{\text{pos}} = \mathbf{E}_{\text{joint}} \oplus \mathbf{P}.
\end{equation}
A learnable cls token \( \mathbf{e}_{\text{cls}} \in \mathbb{R}^d \) is then prepended along the sequence dimension (first dimension) to capture global list context, resulting in the final input to the Transformer:
\begin{equation}
\mathbf{E}_{\text{eval, in}} = \left[ \mathbf{e}_{\text{cls}}; \mathbf{E}_{\text{pos}} \right] \in \mathbb{R}^{(K+1) \times d}.
\end{equation}
Here, `\( ; \)' denotes concatenation along the sequence dimension, making \( \mathbf{E}_{\text{eval, in}} \) a sequence of \( K+1 \) embeddings (1 cls token + \( K \) items) each of dimension \( d \).

This sequence is fed through a Transformer encoder (with \( L_{\text{eval}} \) layers) to model inter-item dependencies and user-aware context. Each layer uses multi-head attention to capture pairwise relationships between all tokens (cls + items) and a feed-forward network to refine individual token features. The Transformer outputs a context-augmented sequence \( \mathbf{E}_{\text{eval, out}} = [\mathbf{y}_{\text{cls}}; \mathbf{y}_{l_1}; \dots; \mathbf{y}_{l_K}] \in \mathbb{R}^{(K+1) \times d} \), where \( \mathbf{y}_{\text{cls}} \) aggregates global list information and \( \mathbf{y}_{l_k} \) encodes context-rich features for the \( k \)-th item.

Two prediction heads process \( \mathbf{E}_{\text{eval, out}} \) to output specific estimates: the pointwise head takes the item-specific embeddings \( [\mathbf{y}_{l_1}, \dots, \mathbf{y}_{l_K}] \) and outputs per-item feedback probabilities \( [\hat{y}_{l_1}, \dots, \hat{y}_{l_k}] \) (e.g., predicted click or like probabilities for item \( l_k \)), while the list-wise head uses the global context embedding \( \mathbf{y}_{\text{cls}} \) to generate a holistic list value estimate \( \hat{y}_{\text{cls}}\) (e.g., predicted total engagement duration or cumulative clicks for the entire list).

\subsubsection{Generator Architecture}
The Generator \( \mathcal{G} \) uses an Encoder-Decoder architecture to transform upstream candidate pools into optimal item lists, with design choices that eliminate spurious order bias and enable step-wise generation.

The Encoder processes the initial candidate pool \( \mathcal{I} = [i_1, i_2, \dots, i_M] \) to produce a context embedding that is invariant to the pool’s initial order. For each candidate \( i_m \), its base embedding \( \mathbf{e}_{i_m}^{\text{base}} \) is concatenated with \( \mathbf{e}_u \) to form \( \mathbf{e}_{i_m, u} = \mathbf{e}_{i_m}^{\text{base}} \oplus \mathbf{e}_u \in \mathbb{R}^d \). These embeddings are stacked into a matrix \( \mathbf{E}_{\mathcal{I}} = [\mathbf{e}_{i_1, u}, \dots, \mathbf{e}_{i_M, u}]^T \in \mathbb{R}^{M \times d} \), which is fed through a lightweight MLP (shared with the Evaluator’s feature extractor) to refine feature interactions:
\begin{equation}
\mathbf{E}_{\mathcal{I},\text{refine}} = \text{ReLU}\left( \mathbf{E}_{\mathcal{I}} \mathbf{W}_{\text{refine}} + \mathbf{b}_{\text{refine}} \right) \in \mathbb{R}^{M \times d}.
\end{equation}
To remove upstream order bias, the Encoder applies Deepset~\cite{zaheer2017deep} sum-pooling over \( \mathbf{E}_{\mathcal{I},\text{refine}} \), aggregating all candidates into a single order-agnostic context embedding that summarizes the pool’s intrinsic quality:
\begin{equation}
\mathbf{c}_{\text{gen}} = \sum_{m=1}^M \mathbf{E}_{\mathcal{I},\text{refine}}[m, :] \in \mathbb{R}^d.
\end{equation}
This \( \mathbf{c}_{\text{gen}} \) serves as the starting signal for the Decoder.

The Decoder generates the final list autoregressively, step-by-step, using a Transformer decoder layer to model dependencies between previously selected items. It starts with the initial context \( \mathbf{c}_{\text{gen}} \) as the first element of its input sequence \( \mathbf{S} = [\mathbf{c}_{\text{gen}}] \). 

At each step \( k \):
\begin{itemize}[leftmargin=*]
\item The current sequence \( \mathbf{S} \), containing embeddings of previously selected items or reasoning tokens concatenated with sinusoidal position embedding $\mathbf{P}$ is fed through the Transformer decoder layer with masked multi-head attention to prevent look-ahead to output a refined step embedding \( \mathbf{z}_k \in \mathbb{R}^d \):
\begin{equation}
    \mathbf{z}_k = \text{TransformerDecoder} \left( \mathbf{S} \oplus \mathbf{P} \right).
\end{equation}
\item This \( \mathbf{z}_k \) guides the selection of the next item from the remaining candidates: similarity scores between \( \mathbf{z}_k \) and each candidate’s refined embedding $\mathbf{e}_{i,\text{refine}}$ from \( \mathbf{E}_{\mathcal{I},\text{refine}} \) are computed via dot product, scaled by a temperature coefficient, and normalized using softmax to yield selection probabilities:
\begin{equation}
p_k^{sel}(i) = \frac{\exp\left( \mathbf{z}_k \cdot \mathbf{e}_{i,\text{refine}} / \tau_k \right)}{\sum_{i' \in \mathcal{I}_{\text{rem}}} \exp\left( \mathbf{z}_k \cdot \mathbf{e}_{i',\text{refine}} / \tau_k \right)} \quad \forall i \in \mathcal{I}_{\text{rem}},
\label{eq:sel}
\end{equation}
where \( \cdot \) denotes the dot product, and \( \tau_k \) dynamically adjusts the exploration-exploitation balance detailed in Section~\ref{sec:temp}, while larger values flatten probabilities for more exploration, smaller values sharpen them for exploitation.
\item If the current step is the reasoning stage, a reasoning token is generated. If it is the recommendation stage, an item is chosen via Thompson sampling during training or greedy selection during inference, its embedding indexed from $\mathbf{E}_{\mathcal{I},\text{refine}}$ is appended to \( \mathbf{S} \), and the process repeats until an item list of length \( K \) is generated.
\end{itemize}

This design ensures the Generator focuses on intrinsic item relationships, while the Transformer decoder enables it to model sequential dependencies in the generated list.

\subsection{Latent Reasoning Mechanism in \ourname}
The latent reasoning mechanism of \ourname~is core-driven by entropy, a key metric we use to measure recommendation task difficulty. Departing from the "reason first, recommend later" paradigm in existing work~\cite{zhang2025reinforced, zhang2025slow, tang2025think}, we propose a "reasoning while recommending" dynamic framework. By making real-time decisions on whether to trigger reasoning at each step of list generation, this framework effectively addresses the high-entropy challenge in the middle of the list—the peak region of decision entropy revealed in Section~\ref{sec:intro}. This design draws on our introductory analysis of latent reasoning: the latent reasoning process itself is accompanied by a continuous drop in entropy. We thus constructed an entropy-based reasoning trigger mechanism to ensure reasoning resources are allocated precisely to high-difficulty stages where the model truly needs them. This balances performance and efficiency by improving recommendation quality without unnecessary computational overhead.

\subsubsection{Entropy-Guided Latent Reasoning}
As shown in Section 1, the entropy of the selection probability distribution \(p_k(i)\) directly quantifies the model’s decision entropy. High entropy indicates significant hesitation in choosing among candidates (corresponding to the middle segment of list generation), while low entropy means the model has formed clear selection preferences (corresponding to the initial and final segments).

At generation step \(k\), the selection probability \(p_k^{ent}(i)\) for each remaining candidate \(i \in \mathcal{I}_{\text{rem}}\) is computed via dot-product similarity, scaling with the fixed base temperature \(\tau_0\), and softmax normalization. This probability distribution is later used to calculate entropy, which determines whether the model enters the reasoning or recommendation stage.

The specific calculation of \(p_k(i)\) is as follows: using the decoder’s step-wise output embedding \(\mathbf{z}_k \in \mathbb{R}^d\) and each candidate’s precomputed refined embedding \(\mathbf{e}_{i,\text{refine}} \in \mathbb{R}^d\), their dot-product similarity is scaled by \(\tau_0\) (fixed base temperature, not dynamically adjusted) before softmax normalization:

\begin{equation}
p_k^{ent}(i) = \frac{\exp\left( (\mathbf{z}_k \cdot \mathbf{e}_{i,\text{refine}}) / \tau_0 \right) }{ \sum_{i' \in \mathcal{I}_{\text{rem}}} \exp\left( (\mathbf{z}_k \cdot \mathbf{e}_{i',\text{refine}}) / \tau_0 \right)} \quad \forall i \in \mathcal{I}_{\text{rem}}.
\end{equation}
With \(p_k^{ent}(i)\) determined, the entropy \(H_k\) (for judging reasoning or recommending stages) at step \(k\) is calculated as:

\begin{equation}
H_k = -\sum_{i \in \mathcal{I}_{\text{rem}}} p_k^{ent}(i) \log p_k^{ent}(i).
\end{equation}

We then compare \(H_k\) with a given threshold \(H_{\text{th}}\). If \(H_k > H_{\text{th}}\) (indicating a high-entropy stage) and the number of consecutively inserted reasoning tokens (REA) has not exceeded the preset maximum reasoning steps per item \(S_{\text{max}}\), we generate and insert a \textbf{context-aware reasoning token} into the generation sequence. The reasoning token is not a static marker but a context-aware embedding aggregated from candidate features, formed through the following process:
\begin{enumerate}[leftmargin=*]
\item The decoder first outputs a preliminary embedding \(\mathbf{z}_k\) based on the current sequence \(\mathbf{S}\) with sinusoidal position embedding $\mathbf{P}$;
\item Using \(\mathbf{z}_k\) as a query, relevance scores with each remaining candidate’s refined embedding are computed: \(s_i = \mathbf{z}_k \cdot \mathbf{e}_{i,\text{refine}}\);
\item These scores are scaled by the reasoning-stage temperature \(\tau_0 \cdot \alpha\) (detailed in Section \ref{sec:temp}) and normalized to get attention weights \(a_i\), which determine how much each candidate contributes to the reasoning token;
\item The final reasoning token embedding is formed by weighted aggregation: \(\mathbf{z}_{\text{REA}} = \sum_{i \in \mathcal{I}_{\text{rem}}} a_i \cdot \mathbf{e}_{i,\text{refine}}\), which encodes diverse inter-candidate relationships.
\end{enumerate}
Inserting \(\mathbf{z}_{\text{REA}}\) into \(\mathbf{S}\) triggers a new decoder iteration, where the model refines its understanding of selected and candidate item relationships using this aggregated embedding, gradually reducing decision entropy. 

Conversely, if \(H_k \leq H_{\text{th}}\) (low-entropy stage) or \(S_{\text{max}}\) is reached, reasoning stops and item selection proceeds. This mechanism ensures reasoning resources are allocated to high-difficulty stages, balancing performance and efficiency.

\subsubsection{Dynamic Temperature for Exploration and Exploitation} \label{sec:temp}
The temperature coefficient \( \tau_k \) is a core parameter that coordinates both reasoning token aggregation at the reasoning stage and item selection in Equation~\eqref{eq:sel} at the recommending stage, balancing exploration and exploitation. We dynamically adjust \(\tau_k\) based on whether the model is in the reasoning or recommending stage, aligning exploration-exploitation strategies with real-time task needs.

In the reasoning stage (when inserting reasoning tokens), we increase the temperature coefficient to \(\tau_{{t}} = \tau_0 \cdot \alpha\) (where \(\tau_0\) is the base temperature and \(\alpha > 1\) is the exploration amplification factor). A larger \(\tau_{{t}}\) flattens the selection probability distribution \(p_k(i)\), prompting the model to explore more potential relationships between candidates during reasoning—critical for breaking decision deadlocks in high-entropy stages and reducing entropy. In the recommending stage, we decrease the temperature coefficient to \(\tau_{{t}} = \tau_0 / \alpha\). A smaller \(\tau_{{t}}\) sharpens the probability distribution, focusing the model on high-confidence candidates. This allows full utilization of reasoning results in low-entropy stages, reducing the probability of suboptimal selections. 

Through such temperature adjustment, \ourname~actively explores to overcome difficulties during reasoning and decisively exploits to lock in high-quality results during recommending, ultimately achieving dynamic balance between exploration and exploitation.
\subsection{End-to-End Training Pipeline}
\ourname’s training follows a two-stage sequential process, where the Evaluator is pre-trained first and then fixed to provide stable supervision for Generator fine-tuning.

\subsubsection{Stage 1: Evaluator Pre-Training}
The Evaluator is trained on historical interaction logs formatted as tuples \( (u, \mathcal{L}, \mathbf{y}_{\text{point}}, y_{\text{list}}) \), where \( \mathcal{L} = [l_1, \dots, l_K] \) denotes an item list, \( \mathbf{y}_{\text{point}} = [y_{1}, \dots, y_{K}] \) represents binary per-item feedback (e.g., \( 1 \) for a click, 0 otherwise), and \( y_{\text{list}} \) is a continuous scalar capturing list-wise utility (e.g., total session duration). Its training objective integrates two complementary losses: a point-wise cross-entropy loss to align with per-item feedback, and weighted log regression loss (WLR) to optimize relative list quality~\cite{covington2016deep}.

The point-wise cross-entropy loss ensures the model learns to predict individual item interactions accurately, defined as:
\begin{equation}
\mathcal{L}_{\text{point}} = -\frac{1}{K} \sum_{k=1}^K \left( y_{k} \log \hat{y}_{k} + (1 - y_{k}) \log (1 - \hat{y}_{k}) \right),
\end{equation}
where \( \hat{y}_{k} \in [0,1] \) is the Evaluator’s predicted probability of feedback for item \( l_k \).

The WLR loss for list-wise overall estimation is defined as:
\begin{equation}
\mathcal{L}_{\text{list}} = - \left(y_{\text{list}} \log \hat{y}_{\text{cls}} + \log (1 - \hat{y}_{\text{cls}})\right),
\end{equation}
where \(y_{\text{list}}\) denotes the continuous list-level utility, and \(\hat{y}_{\text{cls}}\) is the list-level global prediction output by the Evaluator based on the cls token. This loss function enables the direct alignment between the global prediction of the cls token and the true list utility, thereby optimizing the model's ability to evaluate the overall quality of the list. The total loss for the Evaluator is the sum of these two losses:
\begin{equation}
\mathcal{L}_{\text{eval}} = \mathcal{L}_{\text{point}} + \mathcal{L}_{\text{list}}.
\end{equation}

\begin{algorithm}[t]
\caption{EGLR Generator Core Flow}
\label{pscode}
\begin{algorithmic}[1]
\Require User $u$, candidate pool $\mathcal{I}$, pre-trained Evaluator $\mathcal{E}$, 
         target list length $K$, lists per sample $G$, entropy threshold $H_{\text{th}}$, max reasoning steps per item $S_{\text{max}}$, base temperature $\tau_0$, $\alpha$.

\State \# 1. Candidate Pool Encoding
\State $\mathbf{E}_{\mathcal{I},\text{refine}}, \mathbf{c}_{\text{gen}} = \text{Encode}(\mathcal{I}, u)$  % Aggregate pool info (no order bias)

\State \# 2. Generative Re-ranking with Entropy-Guided Reasoning
\For{$g=1$ to $G$}
  \State $\mathcal{L}_g = []$, $\mathcal{I}_{\text{rem}} = \mathcal{I}$, $\mathbf{S} = [\mathbf{c}_{\text{gen}}]$, $\text{REA}_{\text{cnt}}=0$
  \While{$|\mathcal{L}_g| < K$}
    \State $\mathbf{z} = \text{Decode}(\mathbf{S})$  % Decoder output
    \State $p^{\text{ent}} = \text{Prob}(\mathbf{z}, \mathcal{I}_{\text{rem}}, \mathbf{E}_{\mathcal{I},\text{refine}}, \tau_0)$  % Probability for entropy calc
    \State $H = -\sum p^{\text{ent}} \log p^{\text{ent}}$  % Step entropy (entropy)
    
    \If{$H > H_{\text{th}}$ and $\text{REA}_{\text{cnt}} < S_{\text{max}}$}
      \State $p^{\text{rea}} = \text{Prob}(\mathbf{z}, \mathcal{I}_{\text{rem}}, \mathbf{E}_{\mathcal{I},\text{refine}}, \tau_0 * \alpha)$ % Aggregate: compute attention weights via \( \mathbf{z} \) & candidate embeddings, then weighted sum
      \State $\mathbf{z}_{\text{REA}} = \text{Aggregate}(p^{\text{rea}}, \mathbf{E}_{\mathcal{I},\text{refine}})$ % Aggregate: compute attention weights via \( \mathbf{z} \) & candidate embeddings, then weighted sum
      \State $\mathbf{S} \gets [\mathbf{S}; \mathbf{z}_{\text{REA}}]$  % Insert reasoning token
      \State $\text{REA}_{\text{cnt}} \gets \text{REA}_{\text{cnt}} + 1$
    \Else
      \State $p^{\text{sel}} = \text{Prob}(\mathbf{z}, \mathcal{I}_{\text{rem}}, \mathbf{E}_{\mathcal{I},\text{refine}}, \tau_0/\alpha)$  % Selection prob (lower temp)
      \State $l = \text{Sample}(p^{\text{sel}}, \mathcal{I}_{\text{rem}})$  % Sample item
        \State $\mathcal{L}_g \gets [\mathcal{L}_g; l]$, $\mathcal{I}_{\text{rem}} \gets \mathcal{I}_{\text{rem}} \setminus \{l\}$
        \State $\mathbf{S} \gets [\mathbf{S}; \mathbf{E}_{\mathcal{I},\text{refine}}[l,:]]$
      \State $\text{REA}_{\text{cnt}} \gets 0$
    \EndIf
  \EndWhile
\EndFor

\State \# 3. GRPO Update Generator Parameters
\State $r_g = \mathcal{E}(\mathcal{L}_g)$  % Reward from Evaluator (DCG or $\hat{y}_{\text{cls}}$)
\State $\theta \gets \theta + \eta \cdot \nabla_\theta \mathcal{J}(\theta)$ with Equation~\eqref{eq:grpo}  % Gradient update
\end{algorithmic}
\end{algorithm}

\subsubsection{Stage 2: Generator Fine-Tuning with GRPO}
The rewards for the Generator include two types of designs: one is the point-wise reward, i.e., the Discounted Cumulative Gain (DCG) calculated based on the per-item prediction scores \(\hat{y}_{l_k}^{\text{point}}\) from the Evaluator, with the formula given as:
\begin{equation}
\text{DCG}_{\text{eval}} = \sum_{k=1}^K \frac{2^{\hat{y}_{l_k}^{}} - 1}{\log_2(k+1)},
\label{eq:dcg_eval}
\end{equation}
where \(K\) is the length of the list, and \(\log_2(k+1)\) is the position discount factor. The other is the list-wise reward, i.e., the global prediction value \(\hat{y}_{\text{cls}}\) output by the Evaluator.

The Generator’s training data is drawn from the user distribution \( \mathcal{U} \) and candidate pool distribution \( \mathcal{I}_{\text{pool}} \) (i.e., \( u \sim \mathcal{U} \) and \( \mathcal{I} \sim \mathcal{I}_{\text{pool}} \)). During training, for each user-candidate pair \((u, \mathcal{I})\), \( G \) re-ranked lists are sampled via the Generator \(\mathcal{G}(\cdot; \theta)\). The rewards of these \( G \) lists are then standardized to obtain the advantage \(\hat{A}_g\): for the \( g \)-th list \( o_g = [l_1, \dots, l_K] \) (with length \( |o_g| = K \)), the advantage is: 
\[
\hat{A}_g = \frac{r_g - \text{mean}(\{r_1, \dots, r_G\})}{\text{std}(\{r_1, \dots, r_G\}) + \epsilon_{\text{norm}}},
\] 
where \(\epsilon_{\text{norm}} = 10^{-8}\) to avoid division by zero, and \( r_g \) is the DCG$_{\text{eval}}$ or \(\hat{y}_{\text{cls}}\) of \( o_g \).

Based on the full on-policy training paradigm (where the policies for sampling and parameter update are completely synchronized), the on-policy Group Relative Policy Optimization~\cite{guo2025deepseek} (GRPO) gradient objective of the Generator is:
\begin{equation}
\nabla_\theta \mathcal{J}(\theta) = \mathbb{E}_{u \sim \mathcal{U}, \mathcal{I} \sim \mathcal{I}_{\text{pool}}, \{o_g\}_{g=1}^G \sim \mathcal{G}(\cdot; \theta)} \left[ \frac{1}{G} \sum_{g=1}^G \left( \sum_{k=1}^K \log p_{k,g}^{\text{sel}}(l_k) \right) \cdot \hat{A}_g \right],
\label{eq:grpo}
\end{equation}
where \( \mathcal{G}(\cdot; \theta) \) is the Generator that generates lists via step-wise decisions \( \mathcal{G}_1, \dots, \mathcal{G}_K \), \( p_{k,g}^{\text{sel}}(l_k) \) denotes the selection probability of item \( l_k \) at step \( k \) in the \( g \)-th list, and the term \( \sum_{k=1}^K \log p_{k,g}^{\text{sel}}(l_k) \) aggregates the log probabilities of all step-wise decisions in generating \( o_g \) to capture the Generator’s confidence in producing the list.

This objective enables the Generator to optimize the quality of list generation while maintaining stable training by maximizing the expectation of within-group normalized advantages, weighted by the confidence of its step-wise decisions.
\subsection{Pseudocode for \ourname}
The pseudocode~\ref{pscode} outlines the core workflow of the Generator in \ourname, focusing on the key mechanisms: encoding of candidate pools, entropy-guided latent reasoning (via context-aware reasoning tokens) during autoregressive list generation, and policy optimization using GRPO with normalized advantages. Implementation details of network architectures (e.g., encoders/decoders) are omitted to highlight the critical logic flow.

\section{Experiments}

\begin{table}[tbp]
\centering\caption {Dataset Statistics.}
\label {tab:datasets}
\begin {tabular}{lccccc}
\toprule
Dataset & Scenario & \#Users & \#Items & \#Lists & List Len \\
\midrule
Ad & E-commerce & $1$M & $349$K & $483$K & $10$\\
KuaiRand & Short-video & $986$ & $11.6$K & $96.5$K & $10$ \\
\bottomrule
\end{tabular}
\end{table}

\begin{table*}[tbp]
\centering
\small
\caption{Performance comparison on re-ranking task. The best result is bolded and the runner-up is underlined. * indicates that the improvement over the best baseline is statistically significant ($p$-value < $0.05$). }
\label{tab:main}
\begin{tabular}{l l cccccccccc}
\toprule
\multirow{2}{*}{Datasets} & \multirow{2}{*}{Metrics} & \multicolumn{3}{c}{Supervise Learning} & \multicolumn{3}{c}{Reinforcement Learning} & \multicolumn{3}{c}{Latent-Reasoning} & OURS \\
 \cmidrule(lr){3-5} \cmidrule(lr){6-8} \cmidrule(lr){9-11} \cmidrule(lr){12-12}
& & DLCM  & PRM & Seq2Slate & EG-Rerank & CMR & LAST & ReaRec & STREAM-Rec & LatentR$^3$ & \ourname\\
\midrule
\multirow{5}{*}{Ad} 
& MAP@$5$ & $0.6122$ & $0.6129$ & $0.6111$ & $0.6016$ & $0.6022$ & $0.6030$ & $\underline{0.6135}$ & $0.6078$ & $0.6099$ & $\mathbf{0.6155}^*$ \\
& MAP@$10$ & $0.6155$ & $0.6160$ & $0.6140$ & $0.6048$ & $0.6057$ & $0.6065$ & $\underline{0.6161}$ & $0.6115$ & $0.6132$ & $\mathbf{0.6185}^*$ \\
& NDCG@$5$ & $0.6908$ & $0.6912$ & $0.6902$ & $0.6820$ & $0.6831$ & $0.6840$ & $\underline{0.6915}$ & $0.6873$ & $0.6892$ & $\mathbf{0.6939}^*$ \\
& NDCG@$10$ & $0.7042$ & $0.7046$ & $0.7029$ & $0.6962$ & $0.6971$ & $0.6980$ & $\underline{0.7044}$ & $0.7012$ & $0.7028$ & $\mathbf{0.7069}^*$ \\
& Evaluator Score & $0.7120$ & $0.7172$ & $0.7190$ & $0.7249$ & $0.7345$ & $0.7370$ & $0.7217$ & $0.7465$ & $\underline{0.7511}$ & $\mathbf{0.7716}^*$ \\
\midrule
\multirow{5}{*}{KuaiRand} 
& MAP@$5$ & $0.6590$ & $0.6594$ & $0.6518$ & $0.6438$ & $0.6455$ & $0.6472$ & $\underline{0.6603}$ & $0.6547$ & $0.6581$ & $\mathbf{0.6661}^*$ \\
& MAP@$10$ & $0.6290$ & $0.6293$ & $0.6209$ & $0.6120$ & $0.6130$ & $0.6142$ & $\underline{0.6285}$ & $0.6223$ & $0.6257$ & $\mathbf{0.6346}^*$ \\
& NDCG@$5$ & $0.5953$ & $0.5957$ & $0.5880$ & $0.5734$ & $0.5741$ & $0.5767$ & $\underline{0.5952}$ & $0.5891$ & $0.5930$ & $\mathbf{0.6017}^*$ \\
& NDCG@$10$ & $0.7475$ & $0.7480$ & $0.7418$ & $0.7371$ & $0.7382$ & $0.7394$ & $\underline{0.7494}$ & $0.7430$ & $0.7450$ & $\mathbf{0.7526}^*$ \\ 
& Evaluator Score & $2.1600$ & $2.1899$ & $2.1952$ & $2.2600$ & $2.2896$ & $2.2925$ & $2.2558$ & $2.3261$ & $\underline{2.3379}$ & $\mathbf{2.3590}^*$ \\
\bottomrule
\end{tabular}
\end{table*}

We conduct extensive experiments and detailed studies to evaluate the performance of \ourname.
% The code is available at \textcolor{blue}{\url{https://anonymous.4open.science/r/EGLR-01C8}}.

\subsection{Expermental Setting}

\subsubsection{Datasets}
We conducted experiments on two real-world datasets: Ad\footnote{https://tianchi.aliyun.com/dataset/56}, a public e-commerce recommendation dataset, and KuaiRand\footnote{https://kuairand.com/}, a large-scale benchmark from a social media platform for sequential short-video recommendation. The Ad dataset includes $1$ million users, $26$ million ad display/click logs, $8$ user profile features (e.g., id, age, occupation) and $6$ item features (e.g., id, campaign, brand); following preprocessing by LibRerank, user records are converted into ranking lists based on browsing timestamps, with items interacted with within 5 minutes sliced into a single list, resulting in $349$,$404$ items and $483$,$049$ final lists. For KuaiRand, we adopt the $1$K version with irrelevant videos removed, further filtering out items with fewer than $50$ occurrences; the processed dataset contains $986$ users, $11$,$643$ items, and $96$,$532$ interaction sequences with a fixed length of $10$.
Table \ref{tab:datasets} summarizes the key statistics of the two datasets.

\subsubsection{Baselines}
We compare \ourname~against several kinds of baselines for re-ranking:

\noindent \textit{Supervised Learning Baselines}:
\begin{itemize}[leftmargin=*]
\item \textbf{DLCM~\cite{ai2018learning}} employs gated recurrent units (GRUs) to capture global contextual dependencies across recommendation lists.
% \item \textbf{GSF~\cite{ai2019learning}} groups candidate items into overlapping subsets and employs a group-specific scoring function to model item representations within each subset.
% \item \textbf{MiDNN~\cite{zhuang2018globally}} is a deep learning-based listwise ranking model that adheres to the core design of its original formulation.
% \item \textbf{SetRank~\cite{pang2020setrank}} is a Bayesian collaborative ranking method that maximizes the posterior probability of set-wise preference comparisons.
\item \textbf{PRM~\cite{pei2019personalized}} leverages transformer blocks to model mutual interactions between items for listwise recommendation.
\item \textbf{Seq2Slate~\cite{bello2018seq2slate}} formulates re-ranking as a sequence generation task through a sequence-to-sequence architecture, sequentially selecting items via a pointer network.\end{itemize}

\noindent \textit{Reinforcement Learning Baselines}:
\begin{itemize}[leftmargin=*]\item \textbf{EG-Rerank~\cite{huzhang2021aliexpress}} adopts an evaluator-generator paradigm, where its generator produces feasible item permutations and its evaluator assesses the listwise utility of each permutation.
\item \textbf{CMR~\cite{chen2023controllable}} enhances EG-Rerank by integrating deepset and hypernetwork components to boost model expressive power.
\item \textbf{LAST~\cite{wang2024not}} builds on CMR and adaptively refines the generator’s parameters for each individual request during inference.
\end{itemize}

\noindent\textit{Latent-Reasoning Baselines}: It is noted that most of the existing latent-reasoning recommendation works primarily focus on the sequential recommendation or LLM for recommendation~\cite{kang2018self, boka2024survey, zhang2025test}. However, we still incorporate their core latent reasoning into the autoregressive stage of the generative re-ranking models.

\begin{itemize}[leftmargin=*]
    \item \textbf{ReaRec~\cite{tang2025think}} enhances user representations via implicit multi-step reasoning, with reasoning position embeddings and two lightweight methods (ERL, PRL), and we implement PRL variant.
    \item \textbf{STREAM-Rec~\cite{zhang2025slow}} is a slow thinking sequential recommender that addresses one-step inference limitations via a three-stage training framework. 
    \item \textbf{LatentR$^3$~\cite{zhang2025reinforced}} shifts to latent reasoning, uses two-stage training (SFT + modified GRPO-based RL), and optimizes recommendation without CoT data.
\end{itemize}

\subsubsection{Evaluation Metrics}
This study employs two kinds of metrics for experimental evaluation. The first set consists of standard ranking metrics, including Normalized Discounted Cumulative Gain (NDCG) and Mean Average Precision (MAP). These metrics are based on a core assumption: user feedback on items remains unchanged even after the recommendation list has been re-ranked. From the perspective of re-ranking modeling, however, this assumption has limitations in terms of rationality. Nonetheless, it remains a concise experimental evaluation paradigm widely adopted in relevant research.  

To address the challenge that offline data fails to capture accurate feedback under an "unseen distribution", the second metric is defined as the Evaluator Score. Within this metric framework, the Evaluator Score is specified as the reward $\text{DCG}_{\text{eval}}$, computed from the evaluator's predictions in Equation~\eqref{eq:dcg_eval}. This metric effectively captures the practical scenario where re-ranking a recommendation list influences user feedback, and is thus widely recognized as being more consistent with real-world online performance~\cite{huzhang2021aliexpress, wang2024not}.

\subsubsection{Implementation Details}
Our algorithm is implemented in Tensorflow~\cite{developers2022tensorflow} based on the open-source recommendation library libRerank\footnote{https://github.com/LibRerank-Community/LibRerank\label{LibRerank}}. To ensure fair comparison, the batch size, embedding size of each feature, and hidden layer size are set to $128$, $16$, and $64$, respectively. All models are optimized using the Adam~\cite{kingma2014adam} optimizer with a learning rate of $0.0005$. For self-attention models, the number of attention heads is fixed at $8$. For \ourname, we fix the maximum number of latent-reasoning steps per item $S_{{\text{max}}}$ to be selected from \{$\color{blue}{1}$, $2$, $3$\}. The entropy threshold $H_{\text{th}}$ is selected from \{$\color{blue}{0.5}$, $1.0$, $1.5$, $2.0$\}, the base temperature $\tau_0$ is selected from \{$0.3$, $\color{blue}{0.6}$, $1.0$\}, and the adjustment factor $\alpha$ is chosen from \{$1$, $\color{blue}{{2}}$, $5$, $10$\}. $\text{DCG}_{\text{eval}}$ is used as the GRPO's reward signal to align with ranking metrics. For latent-reasoning baselines, we perform a grid search over \{$1$, $\color{blue}{{3}}$, $5$, $10$\} to find the optimal value. The group number for GRPO is fixed at $4$. The number of epochs is fixed at $100$. 
Additionally, we adopt KV-cache to optimize the computational efficiency of the generator’s autoregressive list generation process.

\subsection{Overall Performance}
Table \ref{tab:main} shows that \ourname~ achieves the best performance on both the Ad and KuaiRand datasets and outperforms all three types of baselines in key metrics. Supervised learning baselines update based on real historical user feedback, adapt well to offline data distributions and perform excellently in static metrics such as MAP and NDCG. However, they struggle to generalize to unseen data distributions and have much lower Evaluator Scores than other types of baselines. Reinforcement learning baselines follow the opposite trend. They do not deliberately fit static historical feedback but instead optimize their generalization ability to unseen distributions, thus achieving higher Evaluator Scores that are more aligned with the needs of real-world deployment. Latent-reasoning baselines unlock the multi-step decision-making potential in autoregressive processes. ReaRec integrates latent reasoning into its supervised learning framework, outperforming traditional supervised baselines in static metrics like MAP and NDCG. STREAM-Rec and LatentR$^3$, which are based on reinforcement learning, focus more on generalization ability and achieve better Evaluator Scores. \ourname~ further integrates entropy-guided latent reasoning, exploration-exploitation, dynamic temperature adjustment, and GRPO-based optimization. It boosts synergy between reinforcement learning’s generalization ability and latent-reasoning’s multi-step decision-making advantages in re-ranking, resulting in top performance across both static ranking metrics and Evaluator Score.

\begin{figure}[tbp]  % htbp：控制图片浮动优先级（here, top, bottom, page）
    \centering
    % 插入图片：width=\linewidth 适配单栏宽度，可按需调整为0.9\linewidth等
    \includegraphics[width=0.7\linewidth]{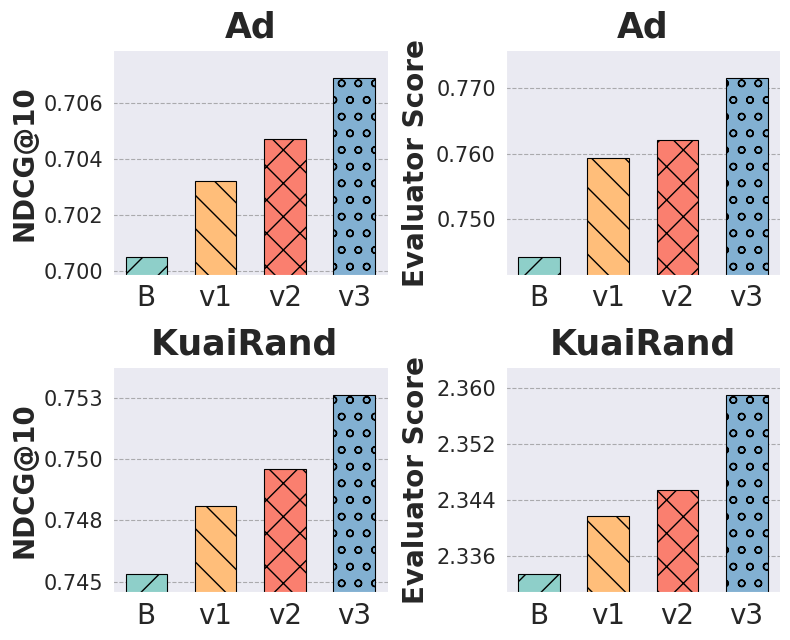}
    % 图表标题：清晰说明内容（数据集、指标、模型对比）
    \caption{Ablation of different model variants (B: backbone generator-evaluator, v1: B + entropy-guided latent reasoning, v2: v1 + context-aware reasoning token, v3: v2 + dynamic temperature).}
    % 标签：用于后文引用（如"见图\ref{fig:model_comparison}"）
    \label{fig:ablation}
\end{figure}
\subsection{Ablation Studies}
To verify the effectiveness of \ourname’s core components, we designed ablation experiments with three model variants and conducted them on the Ad and KuaiRand datasets. The variants include the Backbone (B), which is the backbone generator-evaluator architecture; the v1 variant, which integrates the entropy-guided latent reasoning module into B; the v2 variant, which integrates the context-aware reasoning token into v1; and the v3 variant, which adds a dynamic temperature adjustment mechanism to v2. As shown in Fig. \ref{fig:ablation}, the model’s performance rises steadily as each component is added, proving that both the EGLR module and dynamic temperature mechanism positively improve re-ranking results. Specifically, Entropy-Guided Latent Reasoning resolves uneven entropy distribution in autoregressive generation through entropy monitoring and latent reasoning capabilities. The context-aware reasoning token, by aggregating relationships between selected and candidate items, integrates rich contextual information to strengthen the model’s reasoning process. The dynamic temperature mechanism optimizes the exploration-exploitation balance of reinforcement learning by adjusting temperature coefficients separately during reasoning and recommendation stages.

\begin{table}[tbp]
\centering
\small
\caption{Performance of \ourname~based on different GRMs.}
\label{tab:trans}
\setlength{\tabcolsep}{6pt} % 调整列间距
\begin{tabular}{l c c c c}
\toprule
\multirow{2}{*}{Methods} & \multicolumn{2}{c}{Ad} & \multicolumn{2}{c}{KuaiRand} \\
\cmidrule(lr){2-3} \cmidrule(lr){4-5}
& MAP@10 & E-Score & MAP@10 & E-Score \\
\midrule
Base EG-Rerank & $0.6048$ & $0.7249$ & $0.6120$ & $2.2600$ \\
EG-Rerank$_{\text{EGLR}}$ & $0.6117^*$ & $0.7401^*$ & $0.6182^*$ & $2.3060^*$ \\
\midrule
Base CMR & $0.6057$ & $0.7345$ & $0.6130$ & $2.2896$ \\
CMR$_{\text{EGLR}}$ & $0.6121^*$ & $0.7482^*$ & $0.6191^*$ & $2.3166^*$ \\
\midrule
Base LAST & $0.6065$ & $0.7370$ & $0.6142$ & $2.2925$ \\
LAST$_{\text{EGLR}}$ & $0.6125^*$ & $0.7502^*$ & $0.6199^*$ & $2.3190^*$ \\
\bottomrule
\end{tabular}
\end{table}

\subsection{Cross-Model Transferability of \ourname}
A key core advantage of \ourname~lies in its strong cross-model transferability: its core EGLR module can be seamlessly integrated into various GRMs without requiring extensive architectural modifications or parameter retuning, significantly lowering the threshold for its practical application. This transferability is fully confirmed by the results in Table~\ref{tab:trans}. The three tested backbone models (EG-Rerank, CMR, LAST) differ in their design focuses. For instance, EG-Rerank and CMR emphasize generalization to unseen distributions, while LAST enhances its inference performance during test-time training. Even so, the EGLR module consistently improves their performance across all evaluation metrics on both datasets. However, its performance remains lower than that of LatentR\(^3\) in Table~\ref{tab:main} due to differences in backbone models and non-GRPO optimization methods. This means that \ourname~can be flexibly adapted to various practical re-ranking scenarios where different GRM backbones have already been deployed, providing a flexible and low-cost solution for performance optimization.

% \subsection{Analysis of Latent Reasoning}
% We analyze three key aspects of the latent reasoning mechanism: the impact of maximum reasoning steps ($S_{\text{max}}$) on efficiency-performance balance, the semantic relevance of REA tokens to candidate relationships, and how reasoning refines selection probability distributions. This confirms if the mechanism functions as intended.
\begin{figure}[tbp]
\centering
\includegraphics[width=1.0\linewidth]{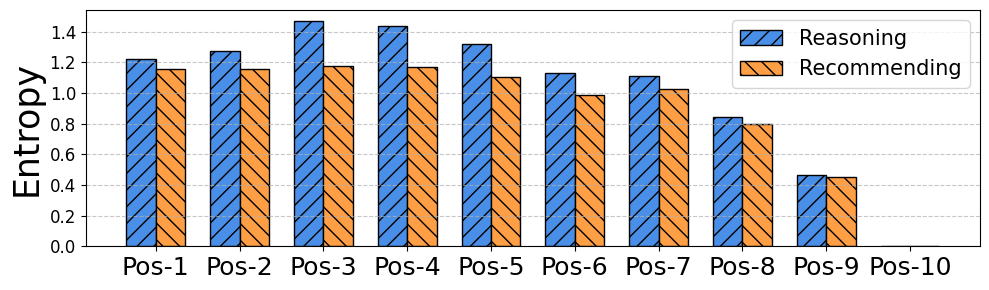} % Replace with actual image path
\caption{Entropy reduction at each position after reasoning.}
\label{fig:ent_ana}
\end{figure}

\subsection{Analysis of Entropy}
We analyzed the entropy variation throughout the entire list generation process, covering both reasoning and recommendation stages. As illustrated in Figure~\ref{fig:ent_ana}, we sampled all reasoning tokens generated at each position during list generation and compared the entropy values before and after reasoning. The results reveal two key findings: first, entropy decreases significantly after reasoning at every position; second, the magnitude of entropy reduction is more pronounced in the middle segment of the list. This phenomenon validates the rationality of using entropy as the criterion for triggering reasoning.

\begin{table}[tbp]
\centering
\caption{Performance and efficiency under different maximum reasoning steps per item on KuaiRand.}
\label{tab:infer_time}
\begin{tabular}{l|cccc}
\toprule
\(S_{\text{max}}\) & $0$ & $1$ & $2$ & $3$ \\
\midrule
MAP@$5$ & $0.6570$ & $0.6661$ & $0.6670$ & $0.6679$\\
MAP@$10$ & $0.6245$ & $0.6346$ & $0.6355$ & $0.6361$\\
NDCG@$5$ & $0.5916$ & $0.6017$ & $0.6025$ & $0.6032$\\
NDCG@$10$ & $0.7443$ & $0.7526$ & $0.7535$ & $0.7541$\\
Evaluator Score & $2.3334$ & $2.3590$ & $2.3602$ & $2.3604$\\
\bottomrule
Inference latency (second) & $38.14$ & $54.89$ & $72.28$ & $86.18$ \\ 
Total reasoning steps per list & $0$ & $7.96$ & $13.5$ & $18.7$ \\
% \\ 
\bottomrule
\end{tabular}
\end{table}

\subsection{Performance and Efficiency of Latent Reasoning Length}
% Since EGLR incorporates a latent reasoning mechanism, it introduces inference latency overhead. To address this, we adopted a KV-cache optimization scheme to offset the latency cost caused by reasoning tokens. We analyzed how KV-cache reduces redundant computations and verified the inference speed of \ourname~to ensure it meets the requirements of real-world deployment. As shown in Table~\ref{tab:infer_time}, we present the inference time and total reasoning steps under different reasoning steps per token \(S_{\text{max}}\), comparing scenarios with and without KV-cache. Given the significant performance improvements brought by \ourname, this computational overhead is acceptable. Furthermore, the model can be adapted to practical deployment by introducing additional reasoning acceleration methods, demonstrating great optimization potential.
As shown in Table~\ref{tab:infer_time}, the performance of \ourname, as reflected by metrics such as MAP, NDCG, and Evaluator Score, exhibits a distinct trend with changes in the maximum reasoning steps per item \(S_{\text{max}}\). Moving from no reasoning (\(S_{\text{max}}=0\)) to a small number of reasoning steps yields noticeable improvements across all performance metrics, indicating that even limited latent reasoning can effectively refine recommendation logic and enhance overall quality. However, as \(S_{\text{max}}\) increases further, the gains in performance become marginal or fluctuate slightly, suggesting that beyond a certain threshold, additional reasoning steps bring diminishing returns in terms of recommendation effectiveness. This pattern highlights that a moderate level of reasoning is sufficient to capture most of the potential performance benefits, while excessive reasoning adds little value to the final results.  

As \(S_{\text{max}}\) increases, inference latency and total reasoning steps per list rise accordingly on a single GeForce RTX 3090 GPU, as more reasoning steps require processing additional tokens, as shown in Table~\ref{tab:infer_time}. However, the efficiency overhead is effectively mitigated by the KV caching technique: this method reuses key and value vectors from past steps, reducing attention computation complexity significantly from \(O(N^2)\) to \(O(N)\) and thereby minimizing redundant calculations. This optimization ensures that the growth in latency remains proportional to the actual increase in reasoning depth, making the trade-off between performance and efficiency feasible for practical deployment.

\begin{table}[tbp]
\centering
\small
\caption{Performance under different numbers of generated item lists (Pass@$K$)
during inference.}
\label{tab:multi}
\setlength{\tabcolsep}{6pt} % 调整列间距
\begin{tabular}{l c c c c}
\toprule
\multirow{2}{*}{Methods} & \multicolumn{2}{c}{Ad} & \multicolumn{2}{c}{KuaiRand} \\
\cmidrule(lr){2-3} \cmidrule(lr){4-5}
& MAP@10 & E-Score & MAP@10 & E-Score \\
\midrule
Pass@$1$ & $0.6185$ & $0.7716$ & $0.6346$ & $2.3590$ \\
Pass@$2$ & $0.6183$ & $0.7718$ & $0.6343$ & $2.3634$ \\
Pass@$4$ & $0.6184$ & $0.7731$ & $0.6346$ & $2.3678$ \\
Pass@$8$ & $0.6181$ & $0.7743$ & $0.6353$ & $2.3727$ \\
Pass@$16$ & $0.6185$ & $0.7753$ & $0.6350$ & $2.3751$ \\
Pass@$32$ & $0.6190$ & $0.7768$ & $0.6347$ & $2.3785$ \\
Pass@$64$ & $0.6194$ & $0.7781$ & $0.6352$ & $2.3810$ \\
Pass@$128$ & $\mathbf{0.6199}$ & $\mathbf{0.7792}$ & $\mathbf{0.6355}$ & $\mathbf{2.3830}$ \\
\bottomrule
\end{tabular}
\end{table}
\subsection{Analysis of Multi-List Generation}
We validated the mainstream industry strategy of generating multiple candidate item lists for each user via beam search and selecting the optimal one using the evaluator score~\cite{wang2024not}. as shown in Table~\ref{tab:multi}, the model performance exhibits a consistent upward trend as the number of generated sequences (denoted by Pass@$K$) increases—this trend holds on both datasets, with no signs of saturation even when $K$ grows to larger scales. Instead, performance continues to show visible improvement, which confirms that the current range of $K$ has not reached a performance bottleneck and proves there remains significant room for optimizing recommendation effectiveness by expanding the scale of candidate lists. Meanwhile, the accuracy improvement already achieved under relatively low $K$ also lays a solid foundation for balancing sampling overhead and recommendation quality in practical applications.

\begin{figure}[tbp]
\centering
\includegraphics[width=1.0\linewidth]{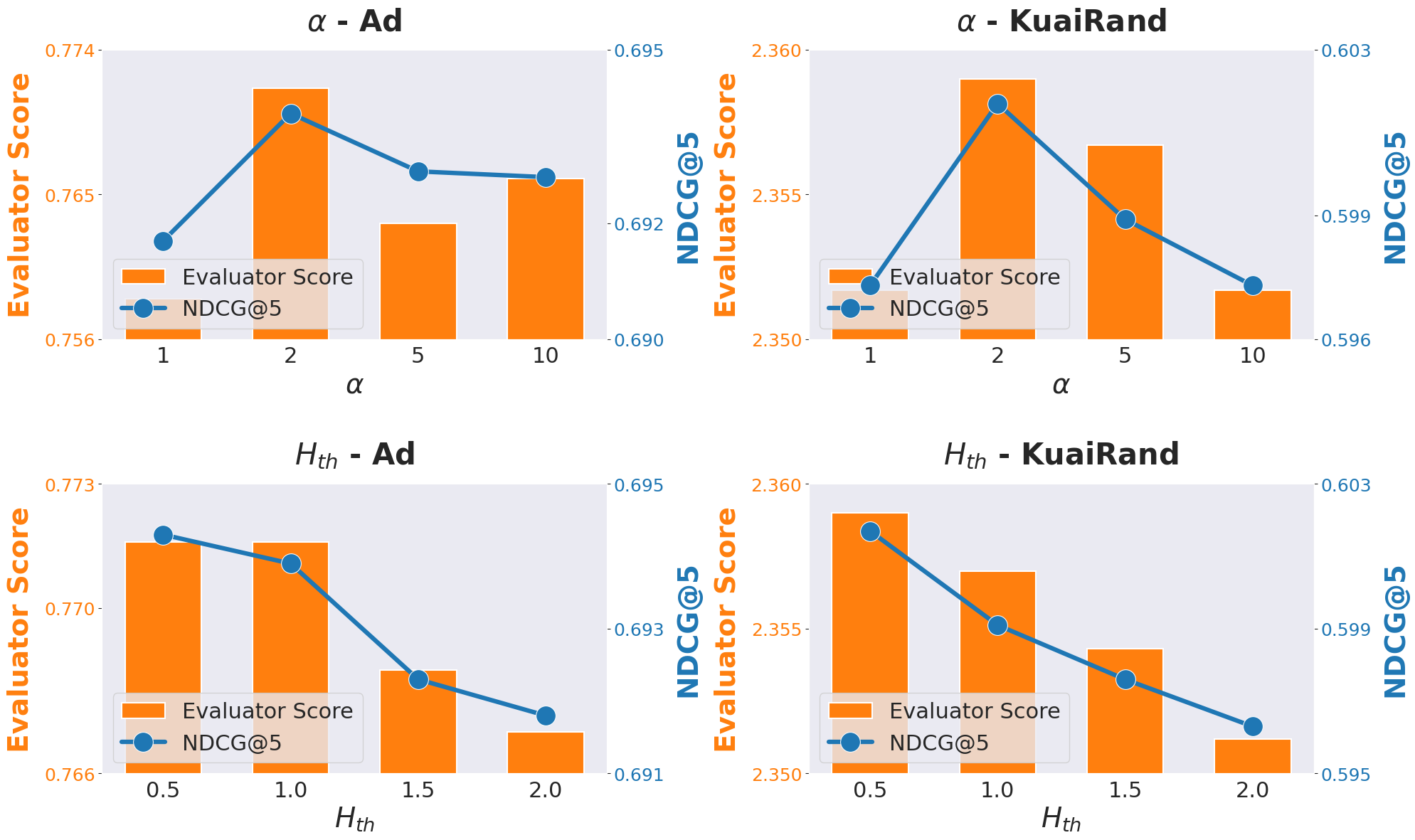} % Replace with actual image path
\caption{Sensitivity Analysis of $\alpha$ and $H_{\text{th}}$.}
\label{fig:ana}
\end{figure}
\subsection{Sensitivity Analysis}

We tested the impact of key hyperparameters on model performance, with results shown in Figure~\ref{fig:ana}. Specifically, the temperature adjustment factor \(\alpha\) controls the uniformity and sharpness of the candidate probability distribution during reasoning and recommendation: when \(\alpha=1\), the factor has no effect on the probability distribution; as \(\alpha\) increases, the entropy of the reasoning process rises while the entropy of the recommendation process decreases. When \(\alpha=3\), the model performance reaches its peak, and further increasing \(\alpha\) does not significantly improve performance, because excessively low temperature may make the reasoning-phase probability distribution uniform, failing to convey effective information.. Regarding the entropy threshold \(H_{\text{th}}\), it serves as the core threshold for controlling the switch between the reasoning and recommendation branches: a larger \(H_{\text{th}}\) enhances the model’s tolerance for high-entropy recommendation results. Conversely, the model needs to increase reasoning steps to reduce recommendation entropy. As shown in Figure~\ref{fig:ana}, as \(H_{\text{th}}\) increases, the model’s reasoning steps gradually decrease, and the performance declines accordingly.

\section{Conclusion}  
% Reinforcement learning is critical for generative re-ranking due to its exploration-exploitation capabilities, but existing generative methods struggle to adapt to dynamic entropy changes during list generation, making it hard to capture complex preferences. Thus, this study draws on the reasoning capabilities of advanced language models to introduce a latent reasoning mechanism, which is experimentally proven to reduce entropy in the decision-making process, and further proposes EGLR, which replaces the "reason first, recommend later" paradigm with "reasoning while recommending" to tackle list generation difficulty, uses context-aware reasoning tokens and dynamic temperature adjustment to realize entropy-guided variable-length reasoning for a better exploration-exploitation trade-off. Experiments on two real-world datasets confirm EGLR’s effectiveness, its ability to boost existing generative re-ranking models, and its practical and research value.

Reinforcement learning is critical for generative re-ranking due to its exploration-exploitation capabilities, but existing methods fail to adapt to dynamic entropy changes in list generation, hindering complex preference capture. Thus, this study introduces a latent reasoning mechanism from advanced language models’ reasoning ability and it is experimentally shown to reduce decision entropy, then proposes \ourname. \ourname~replaces "reason first, recommend later" with "reasoning while recommending" to address list generation difficulty, and uses context-aware tokens and dynamic temperature adjustment for entropy-guided reasoning to balance exploration-exploitation. Experiments on two real datasets confirm \ourname’s effectiveness, ability to boost existing models, and practical as well as research value.

\bibliographystyle{ACM-Reference-Format}
\bibliography{ref}

\end{document}